\documentclass[lettersize,conference]{ieeeconf}

\IEEEoverridecommandlockouts                              
                                                          
\usepackage{cite}
\usepackage{amsmath,amssymb,amsfonts}
\usepackage{algorithmic}
\usepackage{graphicx}
\usepackage{textcomp}
\usepackage{xcolor}
\def\BibTeX{{\rm B\kern-.05em{\sc i\kern-.025em b}\kern-.08em
    T\kern-.1667em\lower.7ex\hbox{E}\kern-.125emX}}
    
\usepackage{xspace}
\let\OldTexttrademark\texttrademark
\renewcommand{\texttrademark}{\OldTexttrademark\xspace }%

\usepackage [english]{babel}
\usepackage [autostyle, english = american]{csquotes}
\MakeOuterQuote{"}
    
\usepackage{hyperref}
\newcommand{\email}[1]{\href{mailto:#1}{#1}}

\usepackage{bm}

\usepackage{cleveref}
\usepackage{afterpage}
\usepackage{gensymb}
\usepackage{paralist}
\usepackage{multirow}
\usepackage{booktabs}
\usepackage{graphicx}
\usepackage{subcaption}  
\usepackage{float}
\usepackage{censor}

\renewcommand{\baselinestretch}{0.99} 

\begin{document}


\title{Quantifying Mechanical Intelligence in Legged Robots with Information Theory}

\author{Zach J. Patterson$^{1}$
\thanks{$^{1}$ Mechanical and Aerospace Engineering, Case Western Reserve University. \email{zpatt@case.edu}}%
\thanks{*This work has been submitted to the IEEE for possible publication. Copyright may be transferred without notice, after which this version may no longer be accessible}%
}

\maketitle

\begin{abstract}
Mechanical intelligence, loosely defined as the reduction in control burden afforded by a robot’s physical form, has become a prominent concept in robotics, with instantiations in bioinspired robotics, soft robotics, robotic swarms, and many other areas. However, rigorous theoretical understanding and quantitative measures of mechanical intelligence have lagged behind the engineering systems that the community has developed. In this work, using modern legged robots as a benchmark and exemplar, we propose several information-theoretic metrics for quantifying mechanical intelligence. By viewing body dynamics as both a computational process and a communication channel, we show that several prior insights in legged-robot engineering can be described using information theory, and we quantify how bits are processed by mechanical modes and across robot coordinates. Specifically, we examine the trade-off between explicitly incorporating compliance through series-elastic actuation and using so-called proprioceptive, low-gear-ratio transmissions, and we explore how these mechanisms interact with control policies during locomotion. We develop these results on systems of increasing complexity: a simplified linear model of a robot-leg transmission, a nonlinear single-leg simulation, and simulated quadruped robots controlled by a learned policy while navigating challenging terrain. These results lay the groundwork for broader study of robot mechanisms and their role in embodied computation.
\end{abstract}

\section{Introduction}

Ever since the early 1980s and the development of hopping robots \cite{raibert1983dynamic,raibert1984hopping}, legged robots have been built to utilize often complex nonlinear dynamics to achieve agile and efficient locomotion behaviors. The underlying perspective, that so-called ``mechanical intelligence'' can be somehow built into the physical design of the robot, has reached into every corner of the robotics enterprise and inspired large sub-communities such as bioinspired robotics and soft robotics \cite{kim2013soft}. Many engineers explicitly see their work in these terms while producing some of the most consequential recent innovations in the field. Nowhere is this more salient than in the work of Dr. Sangbae Kim and colleagues on the proprioceptive actuation concept for the MIT Cheetah platform \cite{bledt2018mit,katz2019mini}, leading directly to the modern boom in affordable and effective quadrupedal and bipedal robots. While Kim and colleagues developed their engineering methodologies to solve the problem of high-rate control for legged robots, they explicitly saw their project as developing ``physical intelligence,'' creating mechanical designs that facilitated easier control solutions \cite{kim2024talk}. Their success in achieving this goal makes these systems ideal benchmarks for understanding mechanical intelligence in general. Specifically, Wensing et al. in their seminal paper \cite{wensing2017proprioceptive} showed that the proprioceptive properties of the low-gear-ratio transmission enable more effective high-bandwidth force control than series-elastic actuator (SEA) transmissions. They performed frequency-domain analysis by linearizing a simplified dynamical model of the system to approximate the open-loop bandwidth properties of various transmission designs. Here, we show that we can reach the same conclusions without specialized analysis, and simply by leveraging general information-theoretic metrics to quantify the mechanical intelligence of the mechanisms. Furthermore, we extend the analysis to simulations of real leg mechanics to show that the methods and results are not limited by the simplifications introduced in the linear model.


Despite the enduring popularity of the mechanical intelligence concept, efforts to formalize and mathematize the field's understanding have lagged behind the engineering innovations. However, there has been some foundational work in this area. In 2004, Pfeifer and Iida provided one of the first descriptions of the topic \cite{pfeifer2004embodied}, and researchers have since sought to apply the concept of reservoir computing to study the information-processing capabilities of soft-bodied robots \cite{nakajima2015information}. There has also been substantial work on using information theory to quantify mechanical intelligence by Ay, Ghazi-Zahedi, and other collaborators \cite{zahedi2013quantifying}. Information theory quantifies the amount of information, defined statistically, that is transmitted or stored during a given process \cite{brillouin2013science,stone2024information}, making it a natural tool to study computation \cite{mackay2003information}. Their approach adopts a particular formulation of mechanical intelligence (or ``morphological computation'', MC, as they refer to it), and they demonstrate that it is useful to describe the sensorimotor behaviors of simple robots operating in mazes \cite{langer2021how}, hopping \cite{ghazi-zahedi2016evaluating}, etc. This literature is the most direct intellectual predecessor of this paper, and the results serve as a critical intellectual foundation. However, that work primarily deals with systems with relatively little mechanical complexity, and thus does not establish a methodology for seeing into the mechanically intelligent system. Existing approaches also fail to scale well to fully passive mechanical systems. This latter issue is critical, as mechanical intelligence at its limit includes systems or subsystems that have no explicit control whatsoever. That said, these works have established the utility of information theory for understanding this foundational robotics problem. 

To further the study of mechanical intelligence in robotic systems, we will present the following contributions. We extend the information-theoretic analysis of mechanical intelligence to systems that are fully passive. We propose several metrics that capture the high-level behavior of mechanically intelligent systems. These metrics are then validated on the MIT Cheetah/Unitree Go family of quadruped robots, which we propose as a benchmark system for this topic. We demonstrate our approach on a linear template model, on a single simulated leg, and on a full quadruped robot simulation controlled with a modern learned policy. Overall, this analysis provides a framework for deeper understanding of mechanical intelligence that can be applied to many other robot systems.

\section{Information Theoretic Analysis}\label{sec:info}

The following is a brief summary of the key information-theoretic results used in this analysis. Given a discrete random variable, $X$, which is distributed according to the probability distribution $p(x)$, we define the entropy of $X$ as $H(X) = -\sum_x p(x) \log p(x)$, which is a measure of the average uncertainty associated with a random variable. The conditional entropy, which measures the average uncertainty associated with a random variable $X$ given that the value of another random variable $Y$ is known, is defined as $H(X|Y) = -\sum_{x,y} p(x,y) \log p(x|y)$.
We can then define the mutual information between two random variables $X$ and $Y$ as 
\begin{equation}
  I(X;Y) = H(X) - H(X|Y) ,
\end{equation}
which measures the amount of information shared between the two variables \cite{shannon1948mathematical}. This mutual information is a natural quantity of interest for analyzing mechanical intelligence, as it allows us to quantify, for example, the number of bits that the body has about the world. 

Mutual information has a number of extensions that allow us to analyze more complex systems. For example, we can define the conditional mutual information between $X$ and $Y$ given a third variable $Z$ as
\begin{equation}
  I(X;Y|Z) = H(X|Z)- H(X|Y,Z) \,,
\end{equation}
which measures the amount of information shared between $X$ and $Y$ that is not already contained in $Z$. This is particularly useful for analyzing systems with multiple interacting components, as it allows us to isolate the information shared between specific components while accounting for the influence of others.


With the information theory background out of the way, we can proceed to describing our more specific framework for analyzing mechanical intelligence. To begin, the systems considered in this paper must include the passive case. In other words, they may not have observations in the control-theoretic sense (so no sensors). While sensors are undoubtedly a critical component for understanding MI, it is logical that MI is prior to and independent of sensing; the body must first respond before something can be sensed, and an intelligent mechanical response may never involve a sensorimotor loop. Therefore, we conclude that a key relationship in MI is between the body's state $X$ and the world state and/or parameters $W$, and the proposed methods will begin with this interaction, before extending to the interaction with control, $A$. This thought process is equivalent to viewing the body as a communication bottleneck between the world and the brain. The amount of information available for the body itself to process, or to pass off to the brain, is therefore constrained by $I(W;X)$, the mutual information between the body and the world. This basic quantity is our starting point. However, a number of other metrics will be considered, especially as we incorporate feedback control into the analysis. 

One possible control-oriented metric is $I(W;X|A)$, which is the conditional mutual information between our world parameters and our state given that we know the action. In other words, this metric can tell us how much information is processed by the body independent of the action, which gives us an initial picture of the relative computation between body and controller.
Similarly, Zahedi and Ay introduced the morphological computation metric $MC_W$, which quantifies the amount of information about the next state of the robot \textit{and} world that is contained in the current state, given the current action \cite{zahedi2013quantifying}. This can be calculated by the conditional mutual information (using a slightly different notation from Zahedi) $I(X_{n+1}, W_{n+1}; X_n, W_n | A_n)$, where $n$ is the discrete time step. The conceptual idea is that, as this value increases, the physical interaction between the robot's body and the world performs more computation to determine its next state.

In the discrete setting, calculating the necessary information quantities is straightforward, but continuous deterministic systems require additional care \cite{penocchio2022information}. In this setting, mutual information becomes finite only after 1) assigning a probability distribution to the unknown environment and 2) introducing a finite resolution on the state, which is a common approach in the literature \cite{metzler2004information}. Together, the environment distribution and state resolution allow the finite information quantities used throughout this analysis. We comment on these assumptions, particularly that of finite state resolution, in the Discussion. For linear systems, these assumptions allow us to derive closed-form expressions by using the expression for the mutual information of a linear Gaussian channel 
\begin{equation}
  I(W;X) \;=\; \tfrac{1}{2}\log\det\!\left( I + \sigma^{-2}\, L  K  L^{\!\top}\right),
  \label{eq:kernel}
\end{equation}
where $ K$ is the covariance of the input distribution, $\sigma^2$ is the variance of the state noise, and $ L$ is the linear map from input to output. 

For nonlinear systems, this straightforward closed-form mutual information is not available, and the mutual information must instead be estimated. While this is associated with a number of difficulties \cite{czyz2023beyond}, we use the state-of-the-art InfoNCE estimator \cite{oord2019representation}. This estimator is based on a contrastive learning approach, which allows us to draw samples of the joint distribution of the variables of interest. Briefly, let $f : \mathbb{R} \times \mathbb{R}^{n_z} \to \mathbb{R}$ be a scoring function, called the \emph{critic}, and let $\{(w^{(i)}, x^{(i)})\}_{i=1}^{N}$ be $N$ independent samples with $w^{(i)}$ a scalar and $x^{(i)}$ a vector. The InfoNCE bound is
\begin{equation}
  \hat I \;=\; \log N \;-\; \frac{1}{N}\sum_{i=1}^{N}
  \left[ -\log \frac{\exp f(w^{(i)},  x^{(i)})}
                    {\sum_{j=1}^{N}\exp f(w^{(i)},  x^{(j)})} \right].
  \label{eq:infonce}
\end{equation}
The bracketed term is the cross-entropy loss $\mathcal{L}$ of a $N$-way classification problem: given a scalar $w^{(i)}$, identify which of the $N$ observations in the batch was produced by it. Therefore, the estimated information is $\hat I = (\log N - \mathcal{L})/\log 2$ in bits. We give more specific information about how we use this estimator for simulation experiments in Section \ref{sec:simulated}.


\section{Analytical Model}\label{sec:proprio}


Following Wensing et al. \cite{wensing2017proprioceptive}, we begin with a simplified model of one leg pressing against the ground. The proprioceptive transmission has one generalized coordinate, $q=\theta$, with inertia $H=m_l r^2+I_{\mathrm{rot}}$. The SEA transmission instead has $q=[\theta_j,\theta_m]^{\!\top}$, with the rotor on an independent motor coordinate $\theta_m$, coupled to the joint coordinate $\theta_j$ by a spring damper of stiffness $k_s$ and damping $b_s$. It has inertia $H=\operatorname{diag}(m_l r^2,I_{\mathrm{rot}})$. Both mechanisms have 
\begin{equation}
	H\ddot{q}+D\dot{q}+Kq
	=J^{\!\top}f(t),
	\qquad
	x_q=\begin{bmatrix}q\\ \dot{q}\end{bmatrix},
	\label{eq:proprio_plant}
\end{equation}
where $J$ maps generalized motion to vertical foot displacement. $J = -r$ for the proprioceptive design and $J = [-r,0]$ for the SEA design. We model the stiffness and damping of the ground, foot, and leg by $k_i$ and $b_i$. For the SEA spring-damper, we define the matrices
\begin{equation}
	 K_{\mathrm{SEA}}=\begin{bmatrix}k_s&-k_s\\-k_s&k_s\end{bmatrix}, D_{\mathrm{SEA}} = \begin{bmatrix}b_s&-b_s\\-b_s&b_s\end{bmatrix}
\end{equation}
These matrices penalize the relative displacement $\theta_j-\theta_m$. The assembled stiffness and damping matrices are
\begin{equation}
	K=K_{\mathrm{SEA}}+J^{\!\top}k_iJ,
	\qquad
	D=D_{\mathrm{SEA}}+J^{\!\top}b_iJ,
\end{equation}
with $K_{\mathrm{SEA}}=D_{\mathrm{SEA}}=0$ for the proprioceptive transmission. We set the series damping to $b_s=2\zeta_s\sqrt{k_sI_{\mathrm{rot}}}$ with $\zeta_s=0.10$. The leg mass, transmission radius, rotor inertia, and contact stiffness are taken from Wensing et al. \cite{wensing2017proprioceptive}: $m_l=3~\mathrm{kg}$, $r=0.1~\mathrm{m}$, $I_{\mathrm{rot}}=0.010~\mathrm{kg\,m^2}$, and $k_i=10^6~\mathrm{N/m}$. Contact damping is set to $b_i=600~\mathrm{N\,s/m}$.

As a check on the model, the proprioceptive transmission has a natural frequency of $79.6$~Hz, compared with the $79.7$~Hz reported in \cite{wensing2017proprioceptive}. The series-elastic design with $k_s=70$~N\,m/rad gives $13.3$~Hz, compared with the reported $13.5$~Hz. As $k_s$ increases, the link and motor coordinates lock together and the series-elastic model converges to the proprioceptive model.


\subsection{Passive system response}
To assess the passive information-processing capabilities of these transmissions, we model the unknown force over a contact window of duration $T$ as band-limited to $B_w$. Such a force has approximately $d=2B_wT$
degrees of freedom over the window. We therefore have a force described by
\begin{equation}
    f(t)=\sum_{j=1}^{d}w_j\varphi_j(t),
    \qquad
    w\sim\mathcal N(0,\sigma_w^2 I),\label{eq:force_passive}
\end{equation}
where $\{\varphi_j\}_{j=1}^d$ are discrete-cosine basis functions spanning frequencies up to $B_w$. We use $T=0.085$~s and $B_w = 200$~Hz, giving $d=34$. 

\begin{figure}[t]
    \centering
    \includegraphics[width=0.47\textwidth]{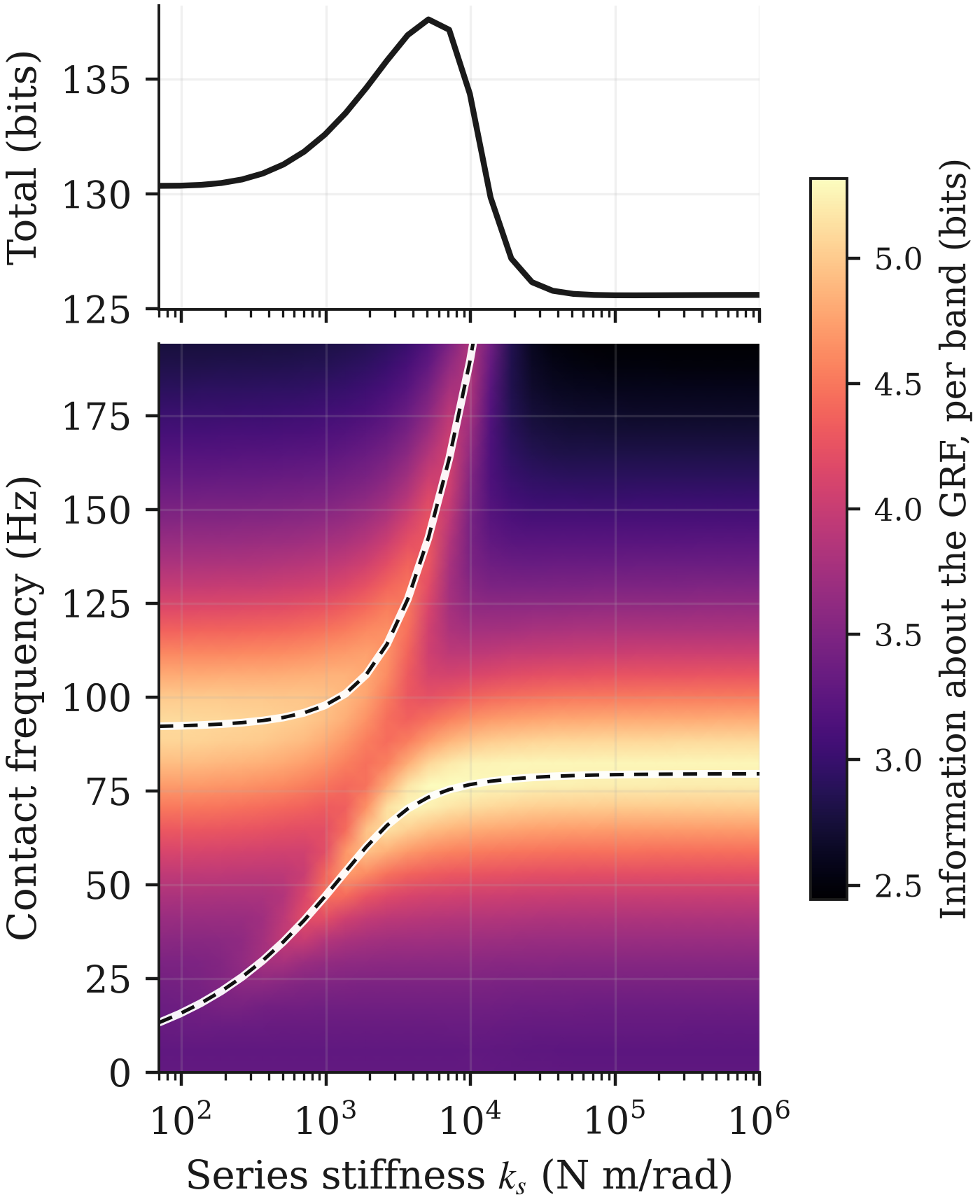}
	\caption{\textbf{Relationship between mutual information and dynamics for the passive leg model.} The simplified robot leg is excited by a sum of randomized forces at different frequencies and amplitudes. The result is swept across series-elastic stiffness values, and the mutual information between the leg state and the force is assessed. \textit{Top:} Total bits present in the passive response. \textit{Bottom:} Bits grouped by their corresponding frequencies. Dashed lines correspond to the natural frequencies of the mechanical modes.}
    \label{passive_spectrum}
\end{figure}

To compare configurations and velocities in a common physical unit and thereby obtain coordinate-invariant noise, we define the energy metric
\begin{equation}
	E=\begin{bmatrix}
		K & 0\\
		0 & H
	\end{bmatrix},
	\qquad x=E^{1/2}x_q.
	\label{eq:energy_whitening}
\end{equation}
The quadratic form $x_q^{\!\top}Ex_q$ measures twice the energy associated with the state. Thus, $x$ expresses every state direction in units of $\sqrt{\mathrm{joules}}$, rather than directly comparing coordinates with different units and inertias. This whitening procedure resembles approaches used in modal analysis \cite{li2007connection}, controllability \cite{moore1981principal}, statistical mechanics \cite{kubo1966fluctuationdissipation}, parameter identification \cite{lee2020geometric}, and, most directly, the Gibbs distribution \cite{jaynes1957information}. We integrate \eqref{eq:proprio_plant} once for each force basis function, apply \eqref{eq:energy_whitening} to whiten the states, and stack the complete whitened trajectories into the columns of a response matrix $M$. The resulting channel is linear and Gaussian, so its information is computed exactly by substituting $L = M$ and $K = \sigma_w^2 I$ into \eqref{eq:kernel}; this can be done per frequency band to produce mutual information at each frequency.


Results of the primary passive system response experiment are shown in Fig. \ref{passive_spectrum}. The top plot shows the total bits of information in the passive response, while the bottom plot shows the frequency-wise distribution of bits. The dashed lines correspond to the natural frequencies of the mechanical modes. As expected, bits are concentrated around the natural frequencies of the system. The total mutual information peaks as the second natural frequency goes to infinity, which implies that the SEA mechanism has more total information available. This is consistent with the common intuition that the more compliant system (with an extra DOF) often has more mechanical intelligence. However, much of the information for the compliant SEA is related to the second natural frequency, which makes accessing this processing capacity during practical control difficult. We will present the implications of this in Section \ref{sec:controlled}.

\begin{figure}[t]
    \centering
    \includegraphics[width=0.47\textwidth]{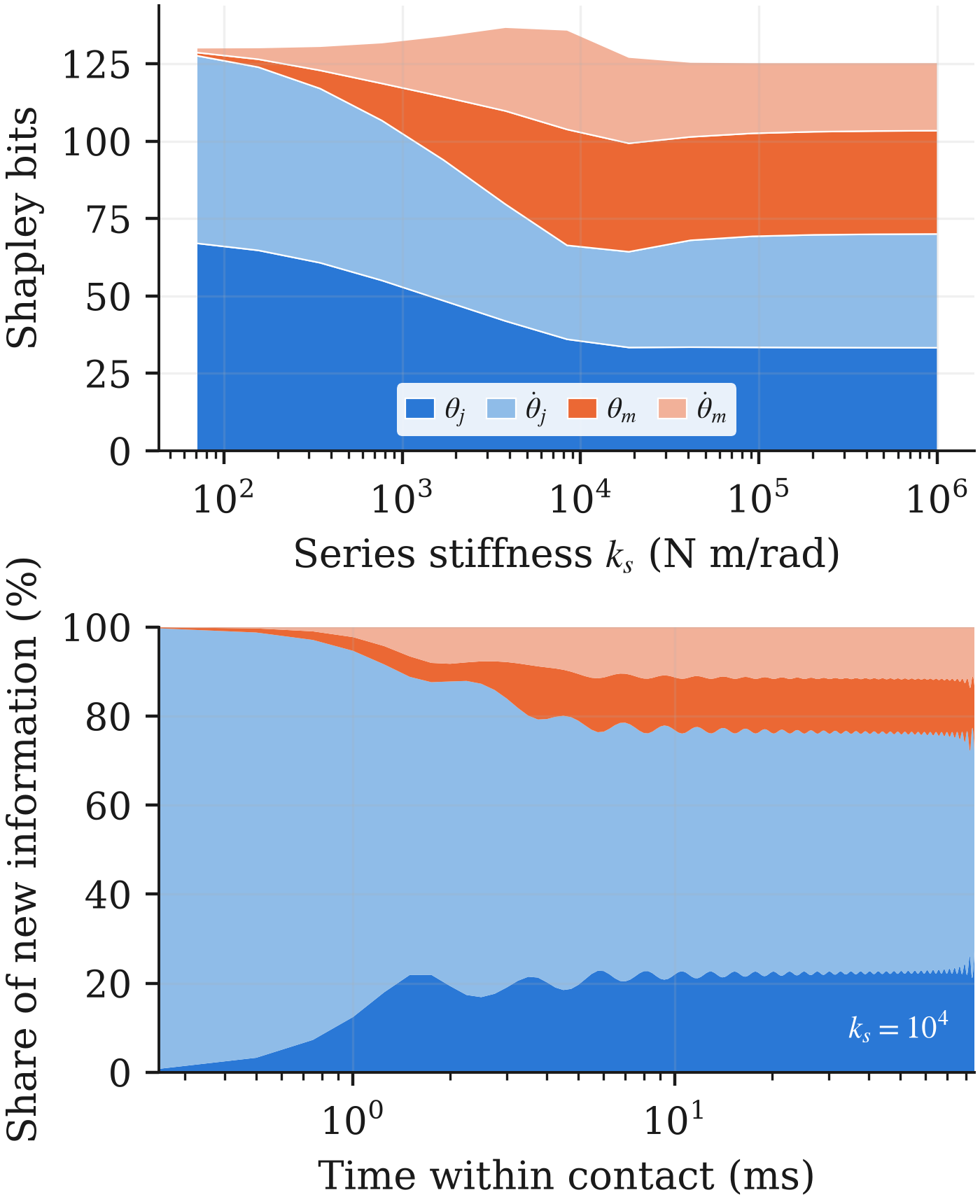}
	\caption{\textbf{Coordinate-wise Shapley share of information content.} \textit{Top:} Share of information contributed by each coordinate over a stiffness sweep. \textit{Bottom:} Temporal evolution of the information share at a single stiffness.}
    \label{passive_coord}
\end{figure}

We also want to examine the information content of each coordinate in the passive response. Because the relative contributions as assessed by the chain rule of mutual information are dependent on ordering, we instead use the Shapley value \cite{shapley1951notes} to assess the contribution of each coordinate. The Shapley value is a game-theoretic concept that fairly distributes the total value of a coalition to its members. For each subset S of coordinates, we compute the information from observing only those coordinates, $\nu(S) = I(W ; X_S)$. Then coordinate $i$'s Shapley value is its chain-rule increment averaged over all orderings:
\begin{equation}
	\phi_i
	= \sum_{S \subseteq N \setminus \{i\}}
	\frac{|S|!\,(n-|S|-1)!}{n!}
	\left[\nu\!\left(S\cup\{i\}\right)-\nu(S)\right].
	\label{eq:coordinate_shapley}
\end{equation}
This rule provides a unique attribution for which the $\phi_i$ sum to the total exactly and coordinates contributing identically to every subset get identical credit.

We show the results of this analysis in Fig. \ref{passive_coord}. The top plot shows the share of information contributed by each coordinate over a stiffness sweep. The bottom plot shows the temporal evolution of the share of information contributed over time for a single stiffness. We see that the SEA's leg-side coordinate contributes a large share of information at low stiffness, but this contribution diminishes as stiffness increases and the two coordinates lock together (these coordinates are redundant at high stiffness, and so give equal Shapley contributions).

\subsection{Simplified controlled analytical model}\label{sec:controlled}

We next want to examine how much computation is performed by the state during feedback control. The controlled plant adds a torque input at the motor, so \eqref{eq:proprio_plant} is modified as follows:
\begin{equation}
	H\ddot{q}+D\dot{q}+Kq
	=J^{\!\top}f(t)+ B_u\tau_d
	\label{eq:controlled_plant}
\end{equation}
where $B_u = [0, 1]^{\!\top}$ for SEA and $B_u=1$ for proprioceptive actuation. The control input, $\tau_d$, is set according to the following control laws. For proprioceptive actuation, we follow Wensing et al. \cite{wensing2017proprioceptive}, and use open-loop force control:
\begin{equation}
	\tau_d= J^{\!\top} f_d.
	\label{eq:proprio_control}
\end{equation}
For the SEA, we instead perform closed-loop control using the spring deflection to measure force, as is standard for SEAs. The control law is
\begin{gather}
	\tau_d = J^{\!\top} f_d + k_p e_f + k_d d_f,\label{eq:sea_control}\\
	e_f = J^{\!\top} f_d - k_s(\theta_j-\theta_m + \nu),
\end{gather}
where $k_p=1$, $k_d=2\zeta\sqrt{I_{\mathrm{rot}}/k_s}$, $d_f$ is the discrete low-pass-filtered (cutoff frequency $400$~Hz) derivative of $e_f$, and $\nu \sim \mathcal N\!\left(0,\ \sigma_\theta^2\right)$. The angular resolution of the force sensor is set as $\sigma_\theta=10^{-5}$~rad. The controller is run at $1$~kHz. For these experiments, the value of $f_d$ has no impact on the information because any known value has no uncertainty associated with it.

We can now compute the total information between the world ($W$) and the body ($X$) and action ($A$) jointly. For these experiments, the world $W$ is the ground reaction force (GRF), $f(t)$, which is parametrized by the randomly drawn amplitude coefficients $w_k$. We then calculate the mutual information conditioned on the action. This quantity measures information uniquely carried by the mechanical state, enabling direct comparison of the mechanical intelligence of the two transmissions. The mutual information is computed as:
\begin{gather}
	I( W; X, A),\\
	I( W; X \mid A)
	=I( W; X, A)-I( W; A).
	\label{eq:controlled_information}
\end{gather}



\begin{figure}[t]
    \centering
    \includegraphics[width=0.47\textwidth]{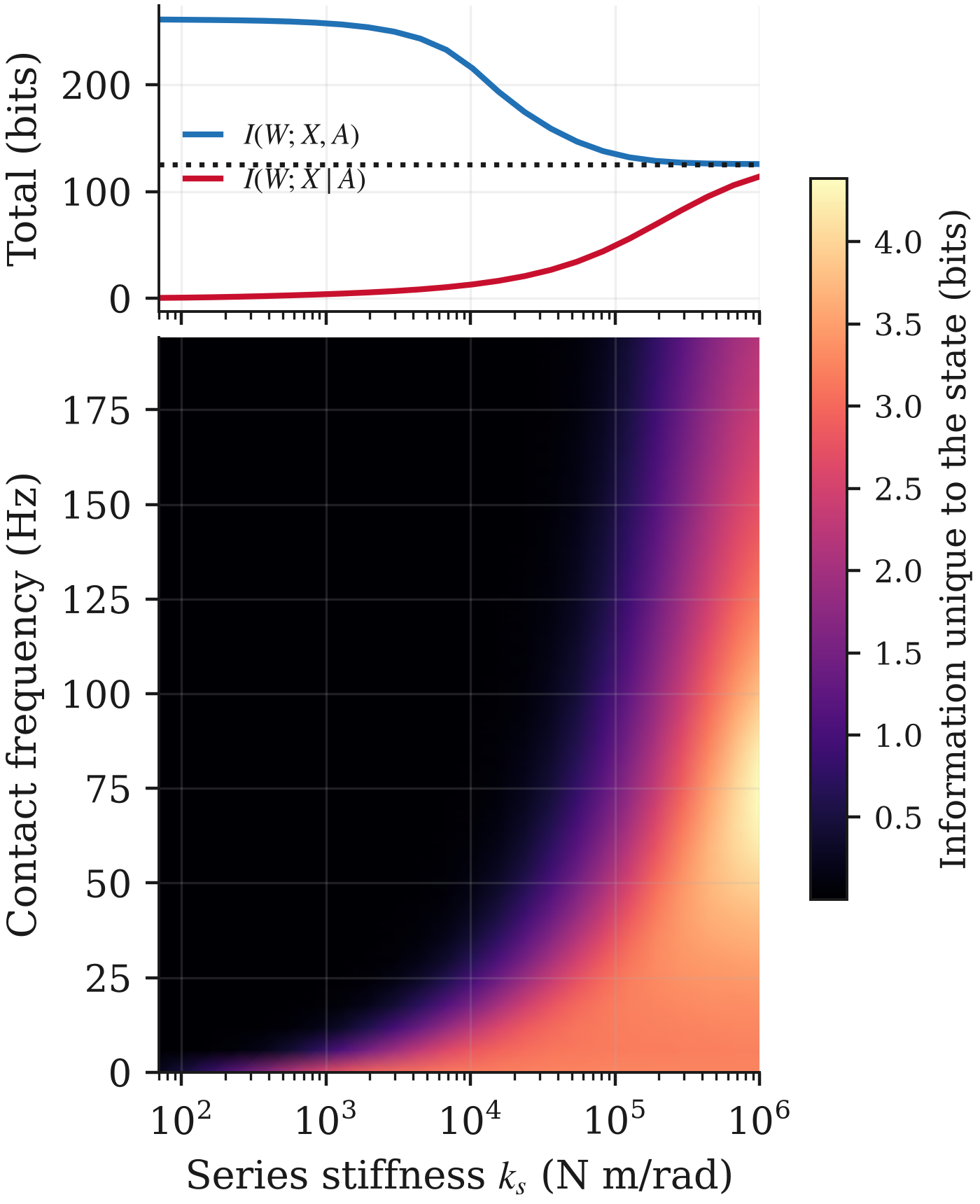}
	\caption{\textbf{Information processing in the body during feedback control.} During a force-control task, the conditional mutual information given the control action \eqref{eq:controlled_information} gives the relative contribution of the body to computation. \textit{Top:} As stiffness increases toward the proprioceptive limit, this value increases even as the total information drops due to the loss of a second mode. \textit{Bottom:} As stiffness increases, the mechanical computation is effective for increasingly higher frequencies.}
    \label{cmi}
\end{figure}
Results of this computation are shown in Fig. \ref{cmi}. While the total information from state and action, $I( W; X, A)$ decreases towards the proprioceptive limit, the conditional mutual information ($I( W; X \mid A)$) increases. This indicates that the stiffer designs accomplish more of the force control task using the body. By looking at the frequency-wise distribution of bits, we also see that stiffer designs perform more mechanical computation at higher frequencies, aligning with the insight that force control is deeply tied to the mechanical bandwidth.


\section{Simulated Robot Leg}\label{sec:simulated}

\begin{figure}[t]
    \centering
    \includegraphics[width=0.47\textwidth]{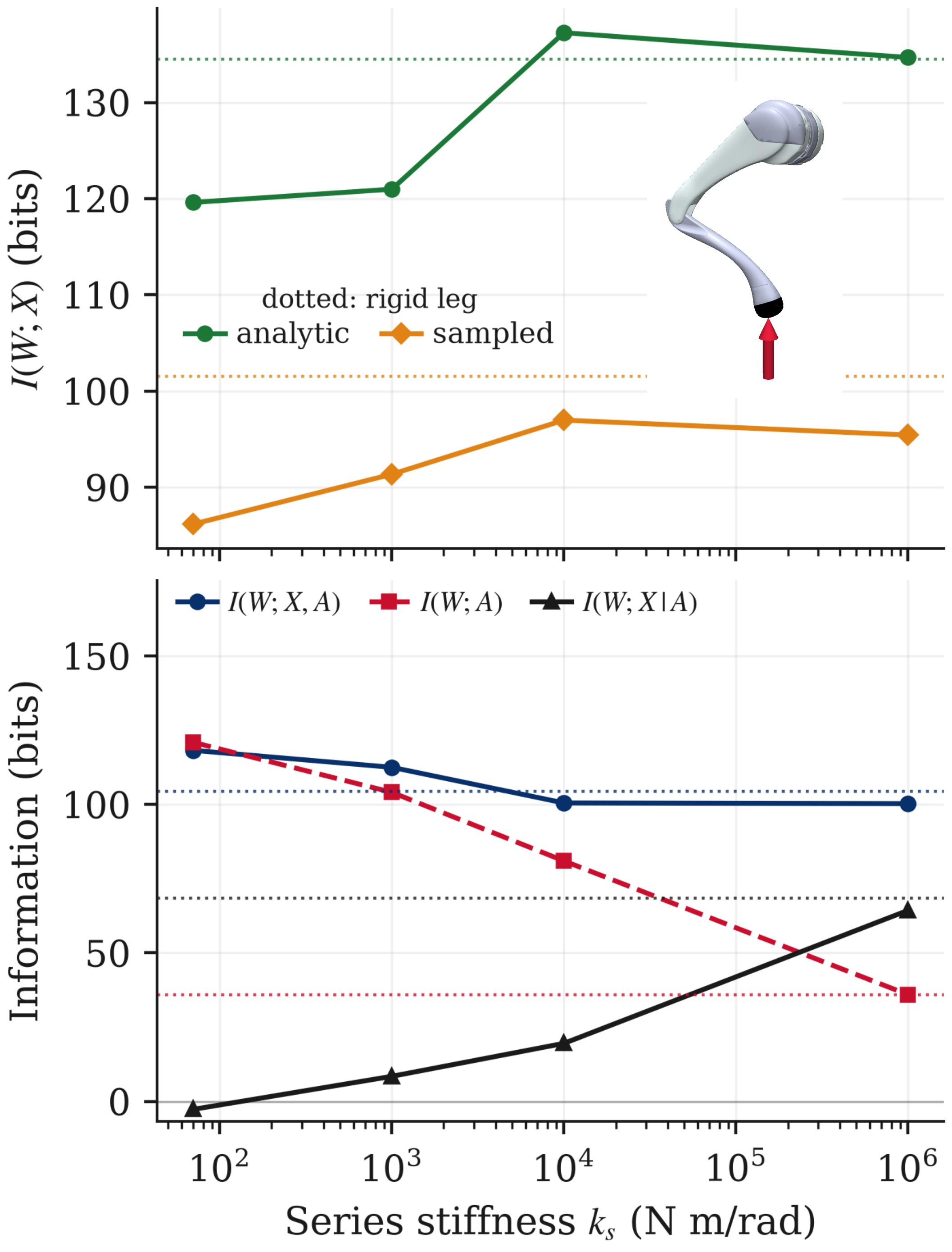}
    \caption{\textbf{Single leg simulation results.} We use the MuJoCo simulator to replicate the results derived from the linear analytic model on a simulated nonlinear robot leg. Again, the goal is to measure how much information about a sinusoidal GRF is present in the state of the leg with and without feedback control. \textit{Top:} Passive force-response experiments, comparing the analytic solution with the sampled nonlinear solution. An image of the simulator is inset. \textit{Bottom:} Feedback-controlled response experiments, with the CMI conditioned on actions produced by force control.}
    \label{leg}
\end{figure}

We perform similar experiments in simulation to those with the analytical model. The simulated robot is the Unitree Go2 quadruped robot. We use the MuJoCo physics engine \cite{todorov2012mujoco} to simulate a single leg of the robot (from MuJoCo Menagerie \cite{menagerie2022github}) with a fixed base (see Fig. \ref{leg} inset for a rendering of the simulation). The leg has 3 degrees of freedom and, for the SEA transmissions, we add an additional hinge joint with a torsional spring to the MuJoCo model at each joint. Each simulation rollout is 85~ms long, the same as in the analytical model, and the timestep is 5e-5~s. We simulate the leg, foot, and ground as a spring-damper acting on the foot, using a boundary condition analogous to that of the linear model. This force is applied as follows:
\begin{equation}
    F_\mathrm{foot} = -k_f (z_f - z_0) - b_f J_z \dot q,
\end{equation}
where $z_f$ is the foot position, $z_0$ is the ground position, $J_z$ is the Jacobian of the foot in the vertical direction. $k_f = 10^6~\mathrm{N/m}$ and $b_f = 2 \zeta_f \sqrt{k_f m_e}$, where $m_e = 1 / J_z H^{-1} J_z^{\top}$ and $\zeta_f = 0.15$. This results in a fixed $b_f = 219.6~\mathrm{N\,s/m}$ for all designs.
We leverage MuJoCo Warp, MuJoCo's massively parallel implementation, to parallelize the simulation across multiple worlds on a GPU, which allows us to run many rollouts in parallel. Each world receives an independent draw of the force coefficients. After simulation, we whiten the state according to the same procedure described in \eqref{eq:energy_whitening}, with the physical terms required to calculate $E$ taken from MuJoCo.

The experiments directly mirror those conducted for the linear model. For the passive experiments, the foot is stimulated with the external force specified by \eqref{eq:force_passive} and N rollouts are collected for each of 5 spring stiffness values. The states are whitened according to the same energy metric \eqref{eq:energy_whitening}, and trajectories are stacked as before. Unlike the linear case, we must use the InfoNCE estimator \eqref{eq:infonce} to calculate mutual informations, which produces a lower bound. We also compare results to the information calculated using the linearization of the nonlinear dynamics. We also perform similar controlled experiments to those performed for the analytic model, using the controllers denoted by \eqref{eq:proprio_control} and \eqref{eq:sea_control}. This time, the commanded force is $f_d = 40\sin(40\pi t) $~N.

We estimate the mutual information as follows. As before, for each frequency band, we estimate the conditional mutual information $I(w_k ; x \mid w_{<k})$, where $w_k$ is the unknown force coefficient of band $k$ from \eqref{eq:force_passive} and $x$ is the whitened and flattened state trajectory. We collect $340$ rollouts per band, giving pairs $(w_k, x)$. To realize the desired conditional mutual information (conditioned on the frequencies below $k$) the coefficients below $k$ are held fixed across all rollouts
used for that term.
InfoNCE requires a critic, which we design as a pair of multilayer perceptrons for each band $k$, in the separable form
\begin{equation}
  f(w, x) \;=\;  g(w)^{\!\top}  h( x),
  \label{eq:separable}
\end{equation}
where $ g : \mathbb{R} \to \mathbb{R}^{32}$ and
$ h : \mathbb{R}^{n_z} \to \mathbb{R}^{32}$ each have two hidden layers of width $256$
with ReLU activations. 
This separable form allows \eqref{eq:separable} to be evaluated using a single $B \times B$ matrix product after two encoder passes, whereas a critic that jointly takes both arguments would require $B^2$ forward passes.
For the controlled experiments, we concatenate the state with the commanded torque in the observation, but we encode the two signals
separately and sum the embeddings,
$f(w, [ x, u]) =  g(w)^{\!\top}\!\left( h_x( x) +  h_u(u)\right)$,
which we found delivered better results than using a single encoder. 

We train the critic on a stream of rollouts as follows. A buffer of $32,768$ rollouts is held on the GPU and each gradient step draws a batch of $1024$ and takes one gradient update step on the loss. Every $128$ steps the buffer is
discarded and refilled from the simulator, preventing overfitting.
After training, a further $16$ batches of $1024$ rollouts are generated and \eqref{eq:infonce} is evaluated on each. 

\begin{figure}[t]
    \centering
    \includegraphics[width=0.47\textwidth]{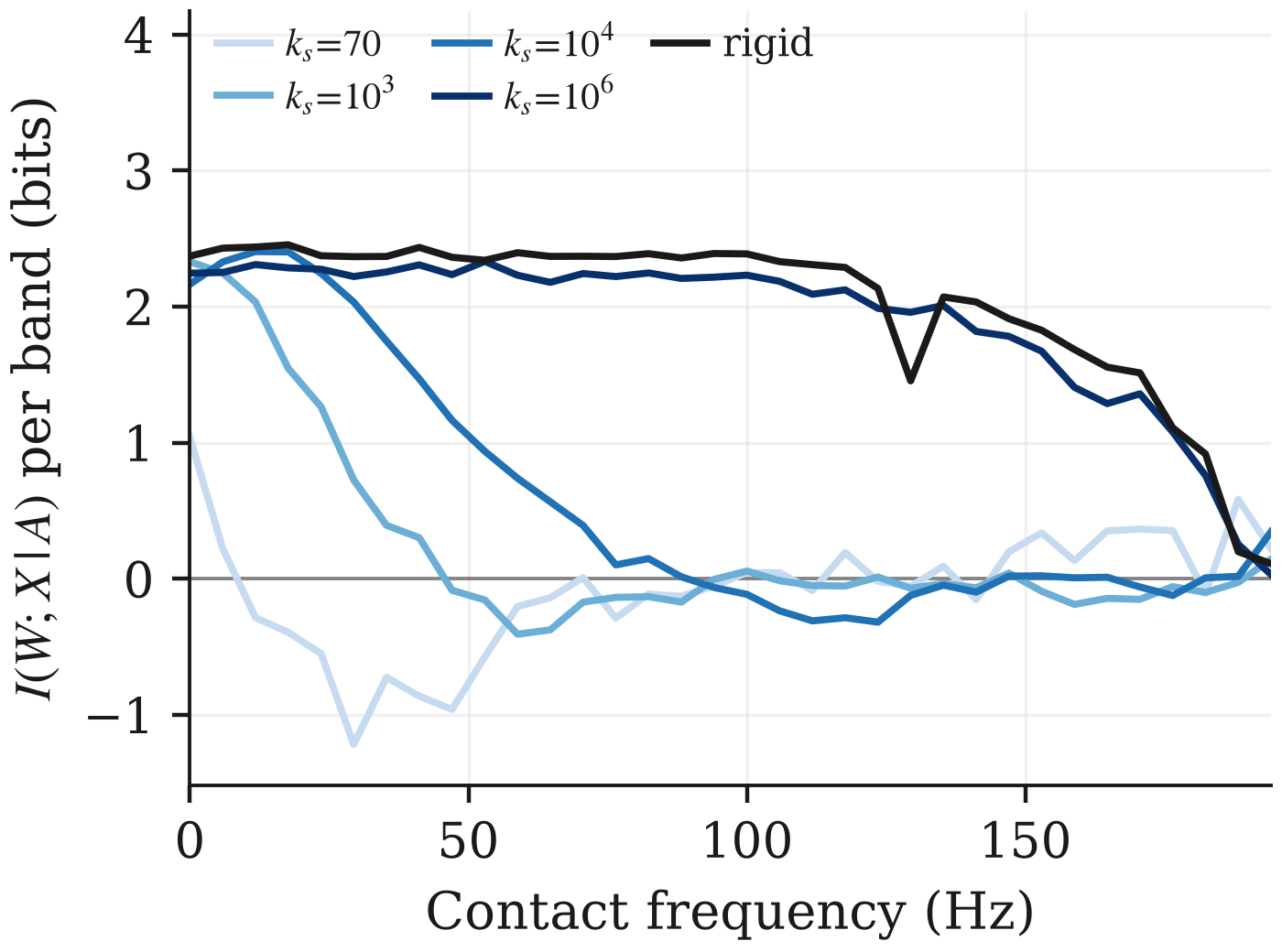}
    \caption{\textbf{Single leg controlled spectra.} Shows the results at individual frequencies corresponding to the total mutual information shown in Fig. \ref{leg}. The modal breakdown of conditional mutual information largely follows the analytic case, with more rigid designs performing more high-frequency physical computation. Negative CMI values are not possible and are the result of estimation error that is more likely when the actual value is close to zero.}
    \label{leg2}
\end{figure}

See Figure \ref{leg} for results from the simulated leg, which is shown in the inset. Passive experiments are shown in the top figure, which shows the information contained in the state of the leg about the GRF. While the simulation estimate is likely an underestimate of the true information due to the lower-bound properties of the InfoNCE estimator, the result shows good qualitative agreement with the analytic results (reproduced at the same points). We see a moderate peak at high stiffness due to the activation of high-frequency modes, followed by a convergence to a steady value. The bottom plot shows the information about GRF during active force control. Results here are simulation-only, and we see a similar trend to that of the analytic results from Fig. \ref{cmi}, with the conditional mutual information $I(W;X\mid A)$ increasing as the stiffness increases towards the proprioceptive limit. This again implies that the body is doing relatively more of the force-control task for the stiffer transmissions. These insights are extended in Figure \ref{leg2}, which shows the spectral information properties of the simulated mechanism. Similar to the analytic case, the stiffer transmissions have high conditional mutual information in their states at high frequencies, again replicating the high-bandwidth force-control arguments that motivate the proprioceptive transmission. Note that, for the more compliant mechanism, the CMI is often below zero, which is mathematically impossible. Here, it is a numerical artifact of the InfoNCE estimator producing a lower bound for a near-zero quantity. 

\section{Full Robot Simulation}
\begin{figure}[t]
    \centering
    \includegraphics[width=0.48\textwidth]{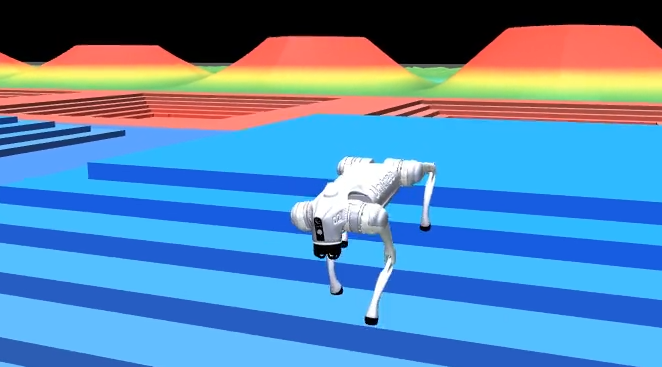}
    \caption{\textbf{Screenshot from velocity tracking environment.} In the velocity tracking task, the robot must learn to navigate challenging terrain while tracking a desired body velocity.}
    \label{sim_screenshot}
\end{figure}
As a preliminary study, we validate our approach on a simulated quadruped robot controlled by a learned policy using reinforcement learning. As before, we use the MuJoCo Menagerie \cite{menagerie2022github} version of the Unitree Go2, which we modify to add series-elastic joints. Because this portion of the paper is meant to be a proof of concept (and due to space constraints), we will present the details at a high level. Our task is velocity tracking, where the robot must simply follow a commanded velocity vector over difficult and varied terrain. We implemented our reward function and training procedures by adapting the Unitree RL Mjlab training procedure for the Go2 \cite{unitreerl2026github}, which in turn is based on the mjlab API for training policies in MuJoCo \cite{zakka2026mjlablightweightframeworkgpuaccelerated}. We train each policy with Proximal Policy Optimization (PPO) \cite{schulman2017proximal} for 3000 iterations.

We experiment with several different robot designs for these simulations, using different spring stiffness and actuator armatures. After training, we collect rollouts to train an InfoNCE estimator according to the same procedure as discussed for the single-leg case. To avoid differences caused by uneven sampling across terrain types, we collect the same number of rollouts on each terrain type. Finally, another difference is that we randomly perturb the robot by applying $15$~N forces on the calves. We call this random force $w_{\mathrm{ext}}$.

\begin{table*}[t]
\centering
\caption{Information metrics across six leg transmissions on the Go2. \emph{n.r.} means not resolvable.}
\label{tab:metrics}
\small
\begin{tabular}{llrrrrr}
\toprule
Design & $k_s$ & Rotor & med.\ error & $I(w_{\mathrm{ext}};x_n\mid x_{<n},u_n)$ & $I(F;x_n\mid x_{<n},u_n)$ & MC$_W$ \\
\midrule
Rigid & inf & 1$\times$ & 0.154 & $0.133 \pm 0.008$ & $1.032 \pm 0.011$ & $0.211 \pm 0.009$ \\
SEA & 70 & 1$\times$ & 0.154 & $0.128 \pm 0.008$ & $0.705 \pm 0.011$ & $0.438 \pm 0.007$ \\
SEA & 300 & 1$\times$ & 0.186 & $0.110 \pm 0.008$ & $0.874 \pm 0.014$ & $0.358 \pm 0.014$ \\
SEA & 300 & 5$\times$ & 0.203 & $0.067 \pm 0.005$ & \emph{n.r.} & $0.300 \pm 0.010$ \\
SEA & 300 & 8.3$\times$ & 0.201 & $0.056 \pm 0.003$ & \emph{n.r.} & $0.303 \pm 0.005$ \\
SEA & 70 & 10$\times$ & 0.205 & $0.044 \pm 0.003$ & $0.404 \pm 0.006$ & $0.273 \pm 0.010$ \\
\bottomrule
\end{tabular}
\end{table*}
We calculate several metrics for the full-body simulation. First, we calculate the median velocity-tracking error over a rollout. The information metric $I(w_{\mathrm{ext}};x_n\mid x_{<n},u_n)$ denotes the information about the disturbance force contained in the state conditioned on the previous state history $x_{<n}$ and the control input $u_n$, where $n$ is the timestep. $I(F;x_n\mid x_{<n},u_n)$ is the same quantity but with $F$ as the ground reaction force, making this quantity similar to what we calculated in other sections of the paper, but conditioned \textit{temporally}. Finally, MC$_W$ is described in Section \ref{sec:info}. Results are shown in Table \ref{tab:metrics}. Generally, we see that performance is negatively correlated with higher rotor inertias, whereas series elasticity has a smaller effect on tracking error, which is a testament to the effectiveness of PPO at finding good control solutions for diverse systems. We also observe that performance is roughly correlated with $I(w_{\mathrm{ext}};x_n\mid x_{<n},u_n)$, which corresponds to a body's instantaneous information processing of a disturbance independent of control. On the other hand, performance is somewhat correlated with $I(F;x_n\mid x_{<n},u_n)$; especially because the ``not resolvable'' values can be thought of as close to zero (they came out negative, which is impossible). This reveals a tension: SEA designs have higher MC$_W$ according to Zahedi's metric, but this metric does not cleanly compare with performance or with the other metrics described in this paper.

\section{Discussion and Conclusions}

We have presented a series of methods for analyzing the mechanical intelligence of two forms of transmissions for legged robots, namely the proprioceptive actuator and the series-elastic actuator. We showed that information-theoretic metrics can readily be used to describe consequential differences in the way that these transmission systems process ground reaction forces through mechanical modes and through their coordinates. Critically, we also presented metrics to describe the amount of computation that is performed by the body independently of the controller. This latter metric aligned with predictions made based on the work of Wensing et al., who showed that proprioceptive actuation should be capable of higher-bandwidth force control than SEAs even in the absence of a sensor to close the loop. This simple concept is therefore an ideal benchmark for developing mechanical intelligence benchmarks, as it is simple enough to admit analytical solutions on simplified versions of the problem while rich enough to force us to deal with the complexity of real robot systems. 

While this work represents progress on the formal analysis of mechanical intelligence, it only scratches the surface of the work to be done on the topic. There are many avenues to strengthen the methods used here to arrive at a richer understanding. First, we would like to expand the work to apply to other types of mechanical systems and robots to solidify the generality and usefulness of the approach. Second, as discussed, the use of information-theoretic methods on deterministic systems is challenging, and further work is needed to formalize the set of acceptable assumptions. In this work, we made pragmatic choices to use stochastic external disturbances and noisy state channels, verifying that the results were sensible and convergent to finite information. However, there is a potential contradiction in the use of the noisy state assumption. It implicitly assumes that mechanical intelligence acts like a sensor, with some finite resolution, as opposed to an infinitely sensitive physical system. This presents a host of philosophical and theoretical issues that will require careful study. Similarly, the energy metric that we use to whiten the state was essential to producing results that were invariant and thus coherent and comparable across different state dimensionalities. The metric has appealing qualities in that it inherently links our information analysis to the physical energy of the system. However, the implications of that link are not tested in this work, and they are deeply related to statistical mechanics and the Gibbs distribution \cite{jaynes1957information}.

Another limitation of the work is the preliminary nature of the full quadruped demonstrations. For a first paper, we considered it important to focus on simpler systems to establish the basic framework, but future work will expand upon our demonstrations of the RL-controlled quadruped and examine other complex robot systems and benchmarks. Of course, the complexity of that setting grows substantially, as we have the interaction of reward function, learning algorithm, robot design, simulation environment, information estimator, etc. Due to the highly statistical nature of many of these components and particularly due to the dependence on stochastic learning, testing across many seeds will be essential to constructing robust conclusions. Finally, while it is not a focus of this work, we have repeatedly found that Zahedi's metric \cite{zahedi2013quantifying}, MC$_W$, does not cleanly relate to those that we have presented. In fact, based on our presented results, it tends to consider SEA designs to be more morphologically intelligent than proprioceptive transmissions. Thus, there is a tension between morphological computation, defined as the body's causal impact on its future state independent of control, and our notion of mechanical intelligence, which is the body's function as a channel between the world and the robot state (and then to the sensors and controllers). We will explore these ideas in future work.

In conclusion, our study replicates key features of quadrupedal robot transmission design principles using information-theoretic metrics. We show how information theory can adjudicate between ``computation'' that is performed by the body and computation that is performed by conventional feedback control. We test these principles on a linear template model, a nonlinear simulated leg, and a full quadruped simulation, demonstrating the scalability and generalization potential of the underlying approach. 

\section{Acknowledgments}
Claude Code was used to assist in software development for this study.



\bibliographystyle{ieeetran}
\bibliography{bib}

\end{document}